\documentclass[10pt,twocolumn,letterpaper]{article}

\usepackage{cvpr}
\usepackage{times}
\usepackage{epsfig}
\usepackage{graphicx}
\usepackage{amsmath}
\usepackage{amssymb}
\usepackage{wrapfig}

\usepackage[pagebackref=true,breaklinks=true,letterpaper=true,colorlinks,bookmarks=false]{hyperref}

\cvprfinalcopy 

\ifcvprfinal\fi
\begin{document}

\title{Recovering 3{D} Planar Arrangements from Videos}

\author{Shuai Du\\
ShanghaiTech University\\
{\tt\small dushuai@shanghaitech.edu.cn}
\and
Youyi Zheng\\
ShanghaiTech University\\
{\tt\small zhengyy@shanghaitech.edu.cn}
}

\maketitle

\begin{abstract}
   Acquiring 3D geometry of real world objects has various applications in 3D digitization, such as navigation and content generation in virtual environments. Image remains one of the most popular media for such visual tasks due to its simplicity of acquisition. Traditional image-based 3D reconstruction approaches heavily exploit point-to-point correspondence among multiple images to estimate camera motion and 3D geometry. Establishing point-to-point correspondence lies at the center of the 3D reconstruction pipeline, which however is easily prone to errors. In this paper, we propose an optimization framework which traces image points using a novel structure-guided dynamic tracking algorithm and estimates both the camera motion and a 3D structure model by enforcing a set of planar constraints. The key to our method is a structure model represented as a set of planes and their arrangements. Constraints derived from the structure model is used both in the correspondence establishment stage and the bundle adjustment stage in our reconstruction pipeline. Experiments show that our algorithm can effectively localize structure correspondence across dense image frames while faithfully reconstructing the camera motion and the underlying structured 3D model.
\end{abstract}

\section{Introduction}
As virtual reality device is becoming more and more popular, 3D content plays a key role associated with these devices. Instead of manually making 3D models using softwares such as 3Ds Max, Maya and Blender, automated 3D reconstruction methods are attracting more attentions due to its high efficiency. By now, a large body of research has been devoted in the field of 3D reconstruction in the aim of producing realistic 3D geometry \cite{furukawa2010towards,furukawa2010accurate,izadi2011kinectfusion,ondruvska2015mobilefusion,snavely2006photo}. However, most of these methods usually output low-level geometric point cloud. These low-level geometry information often lack structure and semantic properties of the 3D content, thus hinders direct usages of these data in the subsequent applications.

In recent, there emerges techniques which exploit high-level structure information such as Manhattan-world assumption \cite{furukawa2009manhattan}, CSG representation \cite{Xiao:2014:RWM}, and repetitions \cite{Ceylan:2014:CSS} to help the process of 3D reconstruction. A common flavour of these approaches is that the structural information such as Manhattan-world assumptions and repetitions are ubiquitous in manmade scenes and thus can be exploited either at the early stage of analysis \cite{Ceylan:2014:CSS} or at the later stage of consolidation \cite{furukawa2009manhattan}.

\begin{figure}
	\centering
	\includegraphics[width=\linewidth]{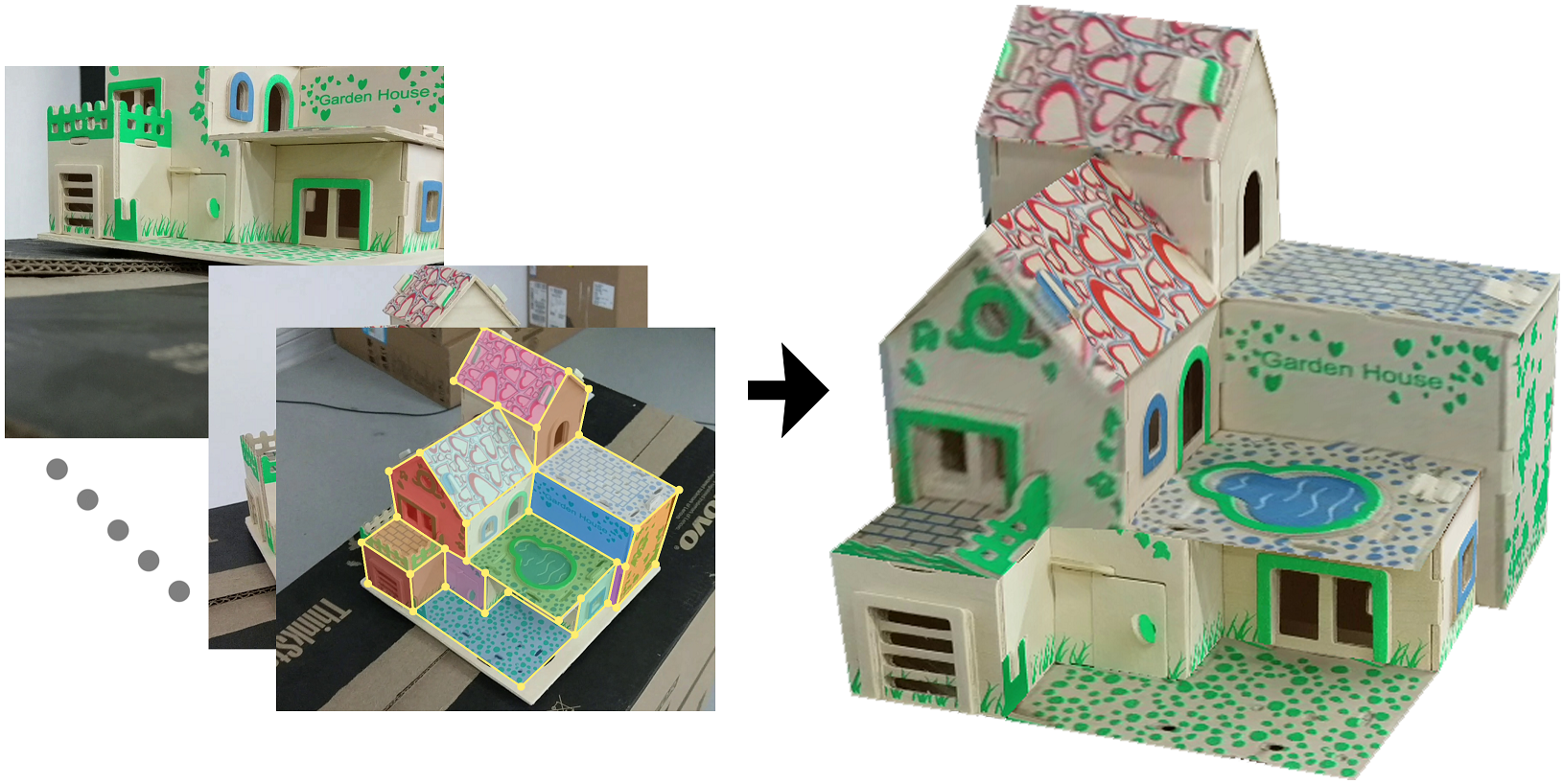}
	\caption{Given an input video sequence of a 3D model and user marked polygons in an initial frame, our method simultaneously recovers a set of 3D planes along with their arrangements which faithfully explain the original 3D model. }
\label{fig:toy_house}
\end{figure}

In this paper, we propose a semi-automatic method to recover the structured 3D model from video sequences. We focus on reconstruction of structured 3D models from videos captured with inexpensive consumer-level RGB cameras. We base our experiments on the speculation that the underlying structure of the 3D object is essentially hidden in the geometry and can be globally decoupled from the geometry \cite{MonszpartEtAl:RAPter:2015}. This could not only give us a more stable bundle estimation but also a structured 3D model. Our key observation is that most manmade environments are constituted of many planar surfaces such as houses and indoor scenes, which inspires us to represent the structured 3D model as a set of planes and their arrangements (e.g., parallel planes and intersecting planes). We devise an optimization framework which simultaneously recovers the plane arrangements and the camera motion as well as the intrinsic relations among the planes.

\begin{figure}[b]
	\centering
	\includegraphics[width=\linewidth]{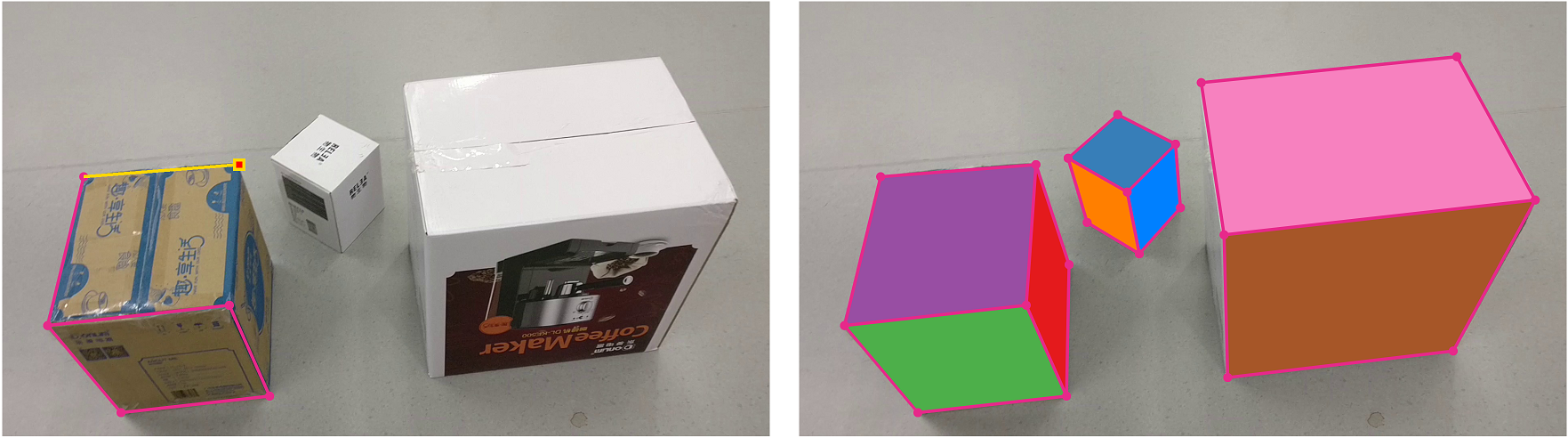}
	\caption{Our user interface allows the user to specify plane points by mouse clicking. Edges are created by connecting previous clicking point and current clicking point. After user specification, we automatically detect the planar faces (shown as colored polygons).}
	\label{fig:UIfigure}
\end{figure}

Start with a video sequence of a 3D model captured with a hand-held consumer camera or downloaded from the internet, we let the user initiate a few planes in one single frame by specifying a sequence of points using mouse clicking (Figure \ref{fig:UIfigure}). We detect planar regions and represent planes in the form of open or closed polygons depending on the point visibility. Our automatic tracking algorithm is then employed to track each polygon point in the rest frames. The tracking consists of a structure-based optical flow process and a backtracking process via dynamic programming. We then use the triangulation algorithm to calculate the motion of the camera and finally, we use a structure-constrained bundle adjustment algorithm to optimize the plane points by taking into consideration of the inter-relations of the planes.

We test our algorithm on various manmade scene data. By manually providing a sparse set of input (points) in the initial frame, we are able to track the structured points in dense image frames using our tracking algorithm. After the tracking procedure, our method automatically extracts structural relations among planes and uses the arrangement relations as constraints in the bundle adjustment. Constraints that fail to be detected by our algorithm can be further added by  the user to provide additional cues to help reconstruct the 3D object. Finally, our algorithm can reconstruct an object satisfying all the constraints.

Our contribution contains two parts. The first part is a structure propagation method. After the user has initially marked out the structure model, we propagate the structure model using a dynamic tracking algorithm, which can relieve the user from marking too many things in order to establish correspondence across the frames. The second part is that we come up with a structure constrained bundle adjustment optimization process, wherein the structural constraints are gradually consolidated during optimization.

We organize this paper's content as follows. Section $2$ discusses the related works in reconstruction and tracking areas. In Section $3$ and $4$, we show the propagation step for dense correspondence matches, the reconstruct algorithm and our modified bundle adjustment method. Results are presented in Section 5. And we conclude with the discussion of our method in Section 6.
		
\begin{figure*}[t]
	\includegraphics[width=\textwidth]{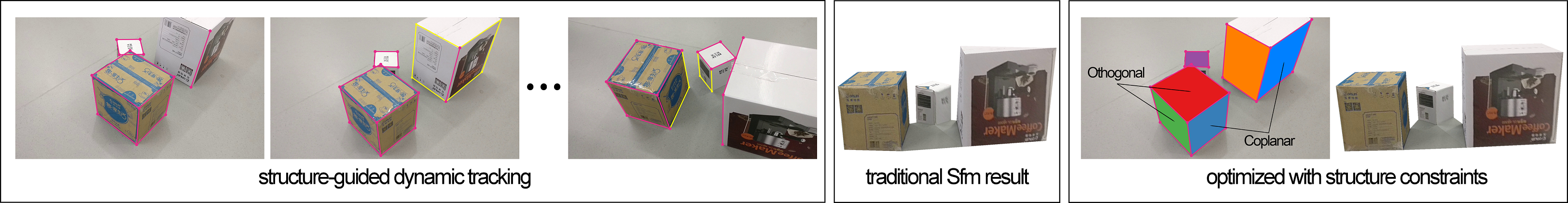}
	\caption{The pipeline of our method. Given the initial marked frame, our method first propagates the structure points across the rest frames by a structure-guided dynamic tracking algorithm, followed with a relation-augmented joint optimization which simultaneously solves for the plane geometry and the plane-to-plane inter-relations (arrangements). Yellow lines are newly added edges which do not appear in previous frames.}
    \label{fig:pipeline}
\end{figure*}

\section{Related Work}
A full review of current state-of-the-art 3D reconstruction algorithms is out of the scope of this paper. We refer interested readers to the excellent surveys on stereo vision \cite{Seitz06acomparison} and multi-view geometry \cite{hartley2003multiple,ma2012invitation}. Below we review the works that are closely related to ours on structure-based tracking and reconstruction.
		
\textbf{Point tracking.} To track the corresponding points between frames, Sam Hare et al. \cite{hare2012efficient} combine matching and tracking together in a unified optimization formulation. They use their method to detect object and track under a large class of 3d pose or homography transformations. We tried a similar version of the method to track the plane corner points using homography warping, but due to the inherent noise in corner point detection which results in the subsequent Ransac computation of homography matrix $H$ unreliable, the tracking result shows to be impractical.
		
Another common way to track for corresponding points is to use optical flow. It is a dense field of displacement vectors which defines the translation of each pixel in a region. Popular techniques for computing dense optical flow include methods by Horn and Schunck \cite{horn1981determining}, Lucas et al. \cite{lucas1981iterative}, and Weinzaepfel et al. \cite{weinzaepfel:hal}.

More recent research works include \cite{wedel2009structure,lempitsky2008fusionflow,brox2011large,bailer2015flow}. Wedel et al. \cite{wedel2009structure} explore fundamental matrix priors which favor flows that are aligned with epipolar lines. Lempitsky et al. \cite{lempitsky2008fusionflow} assume that a number of candidate flow fields have been generated by running standard algorithms possibly multiple times with a number of different parameters. Computing the flow is then posed as choosing which of the set of possible candidates is best at each pixel. And	other methods like Brox et al. \cite{brox2011large} and Bailer et al. \cite{bailer2015flow} first do a coarse feature matching for large displacement optical flow to refine the result.
	
To the best of our knowledge, none of the above methods explicitly exploit a structure model to help the tracking process. Our approach utilizes the underlying planar structure to alleviate the instabilities in single point tracking and thus enables a more reliable frame-to-frame tracking.

\textbf{Structure-based reconstruction.} Many structure-based modeling approaches assume there is a structure. These structures include Manhattan-world assumption \cite{furukawa2009manhattan}, cuboid assumption \cite{jia-3d-stability-pami}, CSG representation \cite{Xiao:2014:RWM}, symmetry \cite{Li07x,Xue2011,jiang2009symmetric} and repetitions \cite{Ceylan:2014:CSS}, etc., which are exploited to help regulate and reconstruct the 3D object and to truly interpret the scene.

By giving pre-known constraints in perspective projection, we can recover the 3D information from a single image \cite{xue2011symmetric,jiang2009symmetric,sturm1999method} by calculating the normals. But single image has a very limited field of view, and can not deal with the occlusion without additional symmetry assumption \cite{jiang2009symmetric,Xue2011}.

Mura et el. \cite{Mura:2015:ARD} use clustered 3D range scans to create the structured 3D models of typical interior environments, namely of recognizing their structure of individual rooms and corridors.

By learning the unique features of different types of surfaces and the contextual relationships between them, Xiong et al. \cite{xiong2013automatic} propose a method to automatically convert the 3D point data from a laser scanner into a compact, semantically rich information model. And from panorama RGBD images, Furukawa et al. \cite{ikehata2015structured} use a graph to	represent the internal structures and reconstruct an indoor scene as a structured model.

Relying on raw outputs of traditional multi-view stereo techniques, a structured model can be created and regularized with structural constraints discovered from the point cloud \cite{xu2016SA,vandenHengel:2007}. Such methods could fail once the multi-view stereo methods return degenerated output due to occlusion, reflectance, and bad illuminations, etc.

In contrast, our method couples the process of structure discovering and structure regularization and jointly optimize the plane geometry and plane arrangements, at the cost of a light-weighted initial input of polygon points.

\section{The approach}
We now detail our algorithm. The main pipeline is devised into two key stages: a structure-based point tracking stage to establish point correspondence across frames and a joint optimization stage where camera motion and planar structures are recovered simultaneously.

\subsection{Initialization of the structure model}
The input to our algorithm is a video sequence of a 3D model or 3D scene captured by hand-hold cameras or downloaded from the internet. As mentioned before, our goal is to reconstruct the structured 3D model in terms of a set of planes and their arrangements. We represent each plane by a planar polygon.

 During initialization, we allow the user to create these polygons manually since automatic detection of planar regions in images is an ill-posed problem and can be easily corrupted by occlusions in cluttered scenes. To create a planar face, the user simply clicks on the image to indicate a corner point of a planar face and then s/he moves the mouse to place the next corner point. Two corner points form a line segment of a planar face. On mouse move, the user sees a highlighted line segment connecting the previous point to the current mouse position (Figure \ref{fig:UIfigure}). In such setting, the user can position the points more precisely by aligning the line segment with image edges. The user continues the task with existing corner points to create additional points and line segments of planar faces. Newly created line segments share vertices with existing line segments. This process leads to a graph with points as graph nodes and line segments as graph edges, for which we use the automatic planar region detection algorithm \cite{Li07x} to extract planar faces. Figure \ref{fig:UIfigure} left shows a snapshot of the interaction process. Since the polygon points are all marked by users, they may be inaccurate. However, this does not affect our point tracking algorithm as our structure-based tracking iteratively improves the points position based on detected structures by minimizing a structure error. This also helps ease users' work in marking the structures at initialization.

Once the user draws up points and lines to form the structured model that needs to be reconstructed, our system automatically tracks these face points using a structured-guided dynamic tracking algorithm. The user can add additional points if they do not appear in the initial frame due to occlusions (yellow points and edges in Figure \ref{fig:pipeline}), our system then automatically tracks the newly added points in the subsequent frames. Occasionally, the user can help the tracking procedure by adjusting the result of some tracking points once occlusion happens or the edges get blurred, and the system will update the intermediate tracking results using dynamic programming. Finally, we will get all the corresponding points to feed into a structure constrained bundle adjustment algorithm.
				
\subsection{Structure-guided Point Tracking}
\begin{figure}[t]
	\centering
	\includegraphics[width=\linewidth]{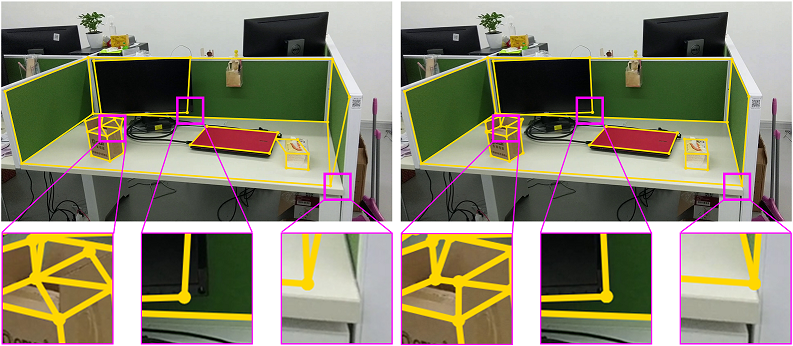}
	\caption{A direct tracking using optical flow leads to undesirable tracking results (left) while our structure-guided dynamic tracking algorithm returns more reliable result (right).}
	\label{fig:trackProcess}
\end{figure}

To trace the corner points of all planar faces in the marked image, a straightforward way is to employ optical flow \cite{lucas1981iterative,weinzaepfel:hal}. Unfortunately, a direct tracking with optical flow returns very bad results as it is based on local gradients without paying any attention to the global structure. See Figure \ref{fig:trackProcess} for an illustration of results from a direct tracing using optical flow.

We resort to an algorithm that exploits the structure information provided in the initial frame. Instead of directly tracing the points in a local window using optical flow, we make the following key observation: a corner point is an intersecting point of its two or more adjacent line segments in the polygonal faces. While tracking of a single point might lead to undesired positions, tracking of a line segment (a set of points) could be more reliable. To this end, we uniformly sample the points along each of the line segments which intersect at a corner point of the planar faces and track the sampled points from the first frame to the next by optical flow. For each line segment $s_i = \{p_1^i,p_2^i,...,p_n^i\}$ where $p_k^i$ is the $k$-th sampled point on $s_i$, each sampled point $p_k^i$ will have a new position ${p_k^{i}}'$, we weight the points with a propagation confidence value computed from the optical flow. We then run a weighted RANSAC algorithm to find the best fitting line associated with the new point positions. Intersection points are updated accordingly which completes the process of corner points tracking. In cases of more than two line intersections, we find the intersection point by weighted least squares. Figure \ref{fig:trackProcess} shows an example of the tracing process. Compared to single-point local tracing, our method generates much more reliable results.

\begin{figure}[t]
			\centering
			\includegraphics[width=\linewidth]{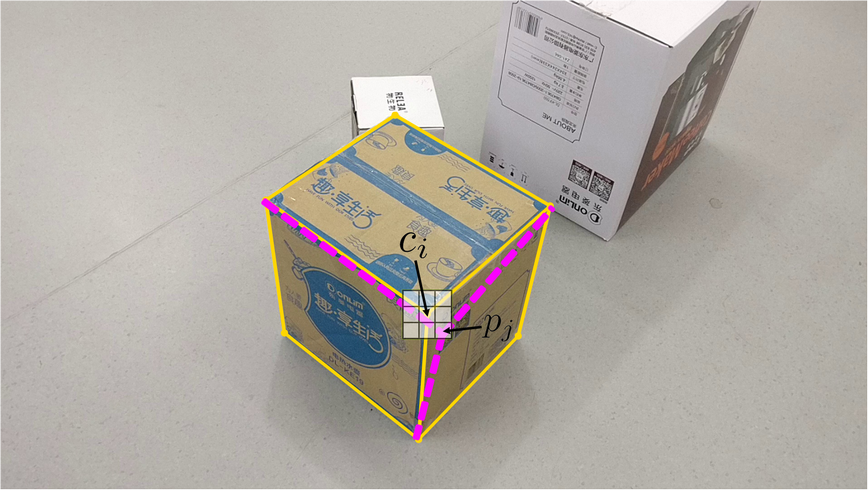} \label{fig:turbulate}
			\caption{To alleviate the influence of noise in the input corner points, for each corner point, we create a local $3\times3$ window and trace all the points in the local window and finally locate the best one as the traced point. The purple line segments are connecting points in the window to the neighboring points of $c_i$.}
\end{figure}

{Sometimes the result of structured optical flow shifts from the real position, this can be caused by a fuzzy point marked by the user. To relieve this problem, we create a local $3\times{3}$ window $w(c_i)$ for each tracking point. We consider each pixel as a candidate point, and we trace all pixels using the structure-guided propagation. In specific, each point $p_j$ in the local window is connected to the neighboring corner points of $c_i$. This creates new line segments (see an illustration in Figure \ref{fig:turbulate}). We then trace these line segments for point $p_j$ using our structure-guided propagation. We choose the most confident traced point (measured as summed weights returned from the optical flow) as the new location for the tracking point of $c_i$. See figure~\ref{fig:turbulate} for an illustration of the process. Each blue line is a candidate edge in the next frame.}

\begin{figure}[b]
	\centering
	\includegraphics[width=\linewidth]{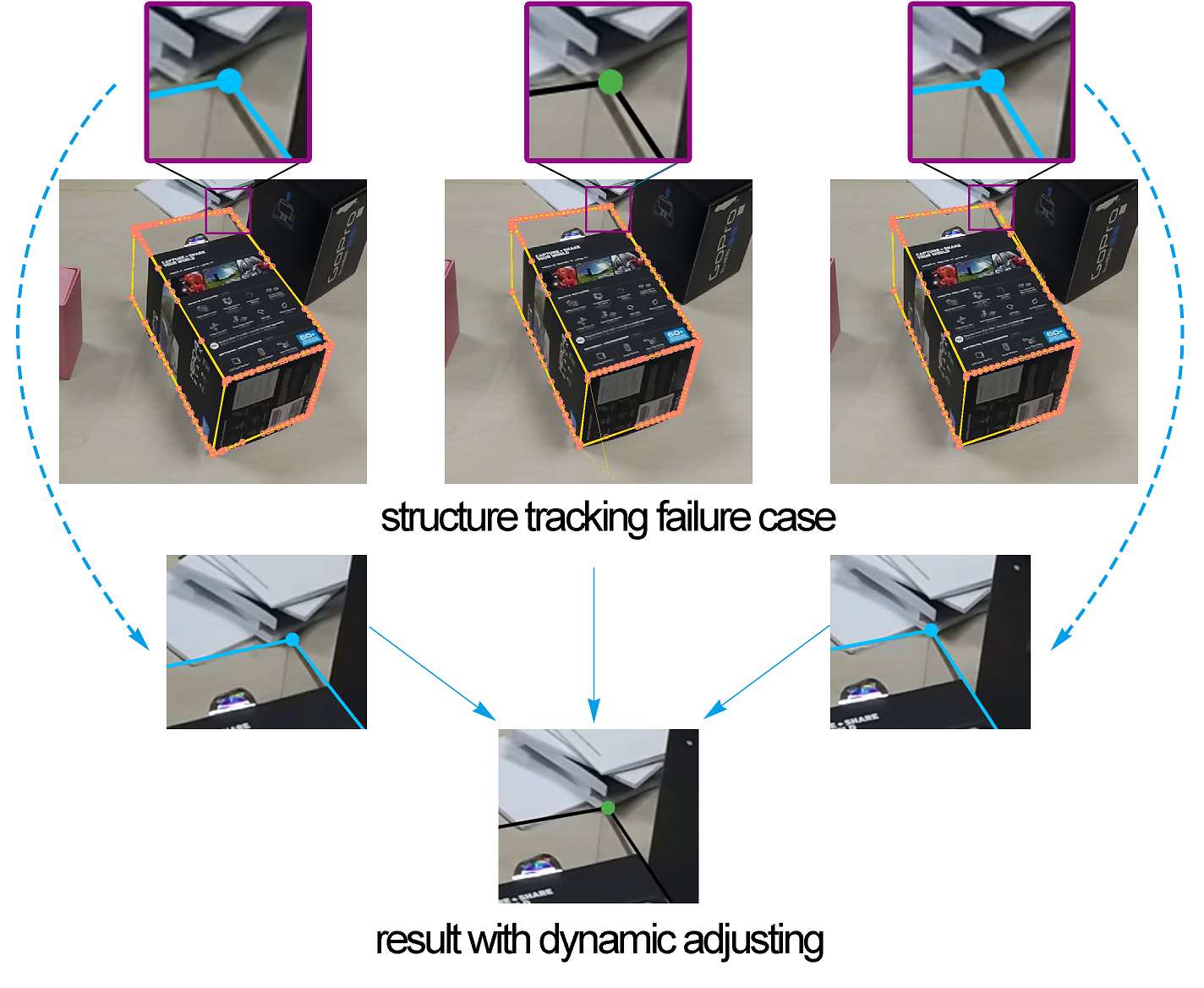}
	\caption{In occasions when occlusion happens or when the edge blends with the background, which can cause our structure-guided point tracking to return inappropriate results. In such cases, the user helps to correct in one frame and the point locations in the intermediate frames are automatically updated using our dynamic back-tracing algorithm.}
	\label{fig:pathFinding}
\end{figure}

Occasionally, the tracing could still fail if occlusion happens or the object edge blends with the background colors as the camera view changes as shown in Figure \ref{fig:pathFinding}. We design a back tracing algorithm to address this issue. Once the tracking is failed in one frame, errors will accumulate in all subsequent frames and it is not invertible since we are not aware of at which frame the tracking goes down. Hence, we leave this to the user. If at any stage the user observes that point tracking goes wrong, s/he could simply adjust the point position by reposition of the corner points using mouse dragging. We devise a dynamic programming to automatically adjust the point positions in all the intermediate frames, detailed below.

Given the start point and the end point in two frames $i$ and $j$, we would like to find a best path connecting these two points and going through points in the intermediate image frames. To increase the possibility of finding the optimal path, as previous, we start from a corner point (pixel) $c_i$ in frame $i$ and create a local $3\times{3}$ window $w(c_i)$ centered at that point. For each point in the window we trace its path along the subsequent frames using the same strategy as mentioned above (i.e., by tracing the line segments). Then each point in window $w(c_i)$ at frame $i$ will be traced to a point in frame $i+1$. In frame $i+1$, we then create for each traced point a local $3\times{3}$ window and repeat the process to frame $i+2$ until we reach frame $j$.

Note that the above process creates a discrete set of local windows across all frames between $i$ and $j$, thus guarantees the existence of a valid tracing path. However, such strategy quickly leads to exponential complexity as the size of local windows grows exponentially. To enable efficiency, we need to bound the search locally such that the windows size does not grow too quickly.

\begin{figure}[t]
	\centering
	\includegraphics[width=\linewidth]{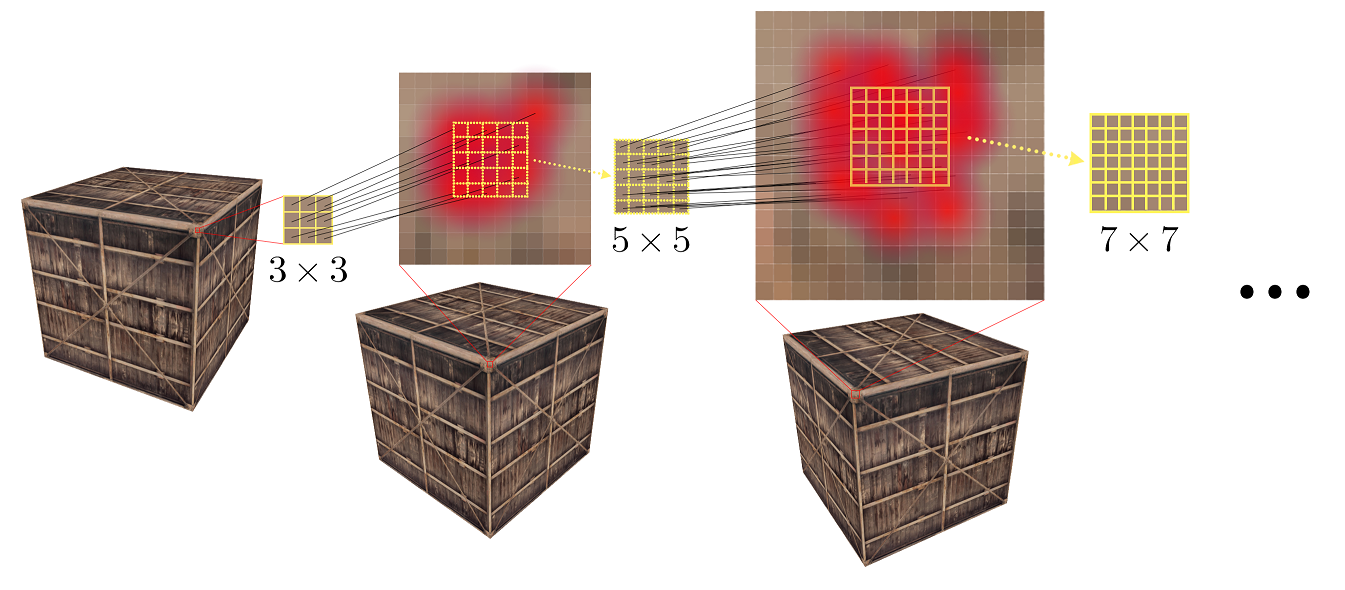}
	\caption{The growth of the tracking windows in our back-tracing algorithm. Each pixel in the previous window propagates to a point in the next frame. The window size increases linearly. The process is then repeated to the next frame until it reaches frame $j$. Each point in a window is linked to all points in the previous and next frames, respectively.}
	\label{fig:trackWindow}
\end{figure}

We devise a constrained window growing algorithm. We observe that the search region should be small when the frame is close to frame $i$ and becomes larger when it is far away from frame $i$. This is not surprising due to the nature of camera motion in consecutive frames. Hence, for each intermediate frame $k$, we restrict local window size to be $\max\{2(k-i)+3, 15\}, k=i,...,j$. The center of local window at a frame $k$ (except for frame $i$ and $j$) is determined as the weighted center of all traced points from frame $k-1$ (weights are computed from optical flow). Note that this will crop out some traced points that are far away from the center. Figure \ref{fig:trackWindow} shows the tracing windows and it can be noted that the positions of window centers vary across frames.

{We then establish links across all intermediate frames.} Each pixel in the local window at frame $k$ is connected to all pixels in the local windows of frame $k-1$ and frame $k+1$ by creating edges. The weight of each edge is the key to our path finding algorithm. We relate it to the results of structure tracking. Let a point $p_i$ denote as the point to trace from frame $i$ and its traced points as $p_{i+1}$ in the next frame. Then the lowest weight is assigned to the link $p_i\rightarrow{p_{i+1}}$ and the weight spreads to the neighbor of $p_{i+1}$ in an increasing manner when connecting $p_i$ to all points in the local window of frame $i+1$, that is, the further it spread, the larger the weight becomes. We used {$(p_{i+1}-p_i)^2 / f_i$ as our weight function, where $f_i$ is a score returned from the optical flow algorithm}. The best tracing path is found as the shortest path connecting points in frame $i$ to frame $j$ using dynamic programming.


				

\section{Reconstruction and Modified Bundle Adjustment}
After tracking, we get the whole sequence of images with corresponding points in them. Following the traditional structure from motion pipeline could give us a set of 3D points as well as the rigid camera transformations. However, this will completely ignore all the planar structures of the model, for example, points of planar faces might not lie on the same plane anymore. Hence, we integrate such constraints in our bundle adjustment algorithm. Besides that, additional structure relations such as coplanarity and orthogonality should be added as well. Detecting such relations from pure 2{d} images is an ill-posed problem due to the lack of 3D information. We thus resort to an iterative optimization approach to analyze such relations from 3D and then re-feed them into the bundle adjustment.

\subsection{Image Formation and Camera Motion}
For the completeness of exposition, we would like to briefly include some basic notions about the camera model we use and a basic description of the structure from motion pipeline. Interested readers are referred to read more sophisticated contents in the excellent book of \cite{hartley2003multiple}.

Assume the camera coordinate of frame $0$ is the world coordinate $\Gamma_w = [r_{1w}, r_{2w}, r_{3w}]$ and the camera coordinate at frame $c$ is $\Gamma_c = [r_{1c}, r_{2c}, r_{3c}]$. Given any 3-D point $X$ in the world coordinate, we have $X=\Gamma_wX_w=\Gamma_cX_c$ where $X_w$ and $X_c$ are the local coordinates of $X$ in $\Gamma_w$ and $\Gamma_c$ respectively (assume $\Gamma_w$ and $\Gamma_c$ share the same origin, otherwise there will be a translation $T_{wc}$). So we have $X_w=R_{wc}X_c$, where $R_{wc}=\Gamma_w^{-1}\Gamma_c$ is a rotation matrix. In general, the coordinates of a 3-D point according to two arbitrary coordinate bases have the following relation:
			\begin{align}
			X_w=R_{wc}X_c+T_{wc},
			\end{align}
	where $T_{wc}$ is the translation between the two corresponding coordinate bases.
			
			We know an image is captured through a camera lens. When the aperture is small, the camera model can be regarded as a pinhole camera. In this case, the point x = $[x,y]^T$ on the image is given by the following equations:
			\begin{equation}
			\begin{aligned}
			x=f\frac{X}{Z},\\
			y=f\frac{Y}{Z}.
			\end{aligned}
			\label{eq:1}
			\end{equation}
			
Here $f$ is the distance between CCD and aperture, $[X,Y,Z]^T$ is the 3D corresponding point of the 2d image. It can be easily derived from similarity geometry.
			
Usually, the image has $(0,0)$ at its up-left corner. It demands a shift in both $x$ and $y$ axis when assuming the focal point lies at image center. Combining all these together, we can get the relation between camera's image plane and world's 3D coordinate:
			\begin{align}
			x' = K[R|t]X'
			\label{eq:3}
			\end{align}
where $X'=[X,Y,Z,1]^T$ and $x'=[x,y,1]^T$ are now homogeneous coordinates, $K = \left[ \begin{array}{cccc} f & 0 & u \\ 0 & f & v \\ 0 & 0 & 1 \\ \end{array} \right]$ is the intrinsic parameter of the camera, $R = R_{cw}$ and $T = T_{cw}$ represent the camera motion related to the world coordinate, and $X'$ is the 3D position under the world coordinate system.

\subsection{Sfm and Optimization}
The relation between two corresponding points $x_1$ and $x_2$ in two images can be derived from equation \ref{eq:3}:
			\begin{align}
			K^{-1}x_1 = RK^{-1}x_2 + T.
			\end{align}
		
			Each pair of R, T can be derived from the eight-point algorithm \cite{hartley2003multiple,ma2012invitation}. After that, we can get relative $R$, $T$ after the eight-point algorithm. We need to merge all relative $R$, $T$ to a world coordinate. Let's say, the relation between the $1^{st}$ and $2^{nd}$ cameras is $[R_{12}|T_{12}]$, the relation between the $2^{st}$ and $3^{nd}$ cameras is $[R_{23}|T_{23}]$. We can easily derive that the relation between the $1^{st}$ and $3^{nd}$ cameras is $[R_{12}[R_{23},T_{23}]|T_{12}]$.

Finally, we run the optimize process to reduce the error. We consider both the re-projection error and the structure error during the optimization. In detail, the re-projection error is formed as:
			\begin{align}
			\min{\sum_i{\sum_j{(x_i^j-K[R_i|t_i]X^j)}^2}},
			\end{align}
			
			where $x_i^j$ is the 2d image point of 3D $X^j$ in image $i$.
			
			Besides the re-projection error, we add an additional term to measure the structure error, which is the points on a same planar face should stay coplanar in 3D. This leads to the constraint:
            \begin{align}
			e_{ij}^i * N_i = 0,
			\end{align}
where $e_{ij}^i$ is an edge on plane $i$, $N_i$ is the normal direction of plane $i$ which can be computed from the edges. Optimizing the above equations leads us to a bundle adjustment of camera motion and structured planar faces. Still, there are other structure relations missing, such as coplanarity between two planar faces and orthogonality between two planar faces.

\begin{figure}[b]
		\centering
		\includegraphics[width=\linewidth]{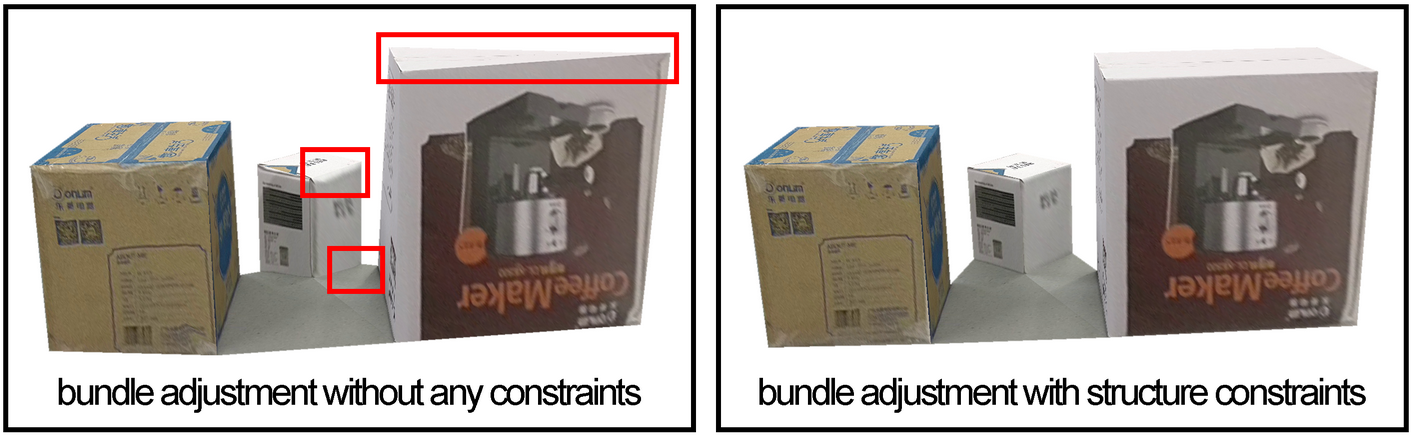}
		\caption{A comparison of results obtained using Bundle adjustment without structure constraints and with structure constraints.}
	\end{figure}

A direct analysis of such relations from 2d is not feasible, thus we detect the coplanarity and orthogonality in the estimated 3{D} from the above bundle adjustment. We employ a method similar to the method of GlobFit \cite{li_globFit_sigg11}. We detect near orthogonal, coplanar, and parallel plane groups and attempt to enforce them to be orthogonal, coplanar, and parallel. A group of planar faces are detected to be parallel if their normals coincide ($\leq 10^\circ)$. A group of planar faces are detected to be coplanar if their normals coincide and the line connecting their centers is close to orthogonal to their normals. Two parallel groups of planar faces are detected to be orthogonal if the angle between their normals is close to $90^{\circ}(\pm{10^{\circ}})$. To reduce the ambiguities in detecting these constraints, we first detect parallel groups of planes and take their weighted normal for the subsequent analysis of orthogonality and coplanarity. To detect parallel groups, we apply mean shift \cite{cheng1995mean} on plane normals with a default bandwidth set to $1e-3$. During the optimization, if any of the group enforcement leads to an increase of error in the bundle adjustment, we release such group constraint. If the automatic detection fails, we let the user to indicate planar relations by clicking on relevant faces. The constraints of orthogonality, parallelism, and coplanarity are in the following forms:
			\begin{align}
			N_a \cdot N_b = 0;\\
            N_i \cdot N_j = 1;\\			
            e_{ij}\cdot(N_i+N_j) = 0.
			\end{align}
			
Here $e_{ij}$-s are edges connecting points in the two planes. We use Levenberg-Marquardt Algorithm to optimize the sparse camera motion.

	\section{Experiments}
	
	\begin{figure}[t]
		\centering
		\includegraphics[width=\linewidth]{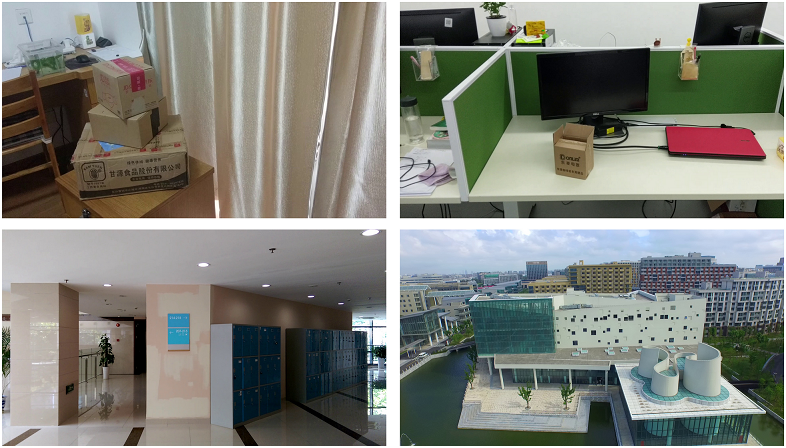}
		\caption{Exemplar scenes used in our experiments.}
        \label{fig:res_example}
	\end{figure}
	
	\begin{figure}[b]
		\centering
		\includegraphics[width=\linewidth]{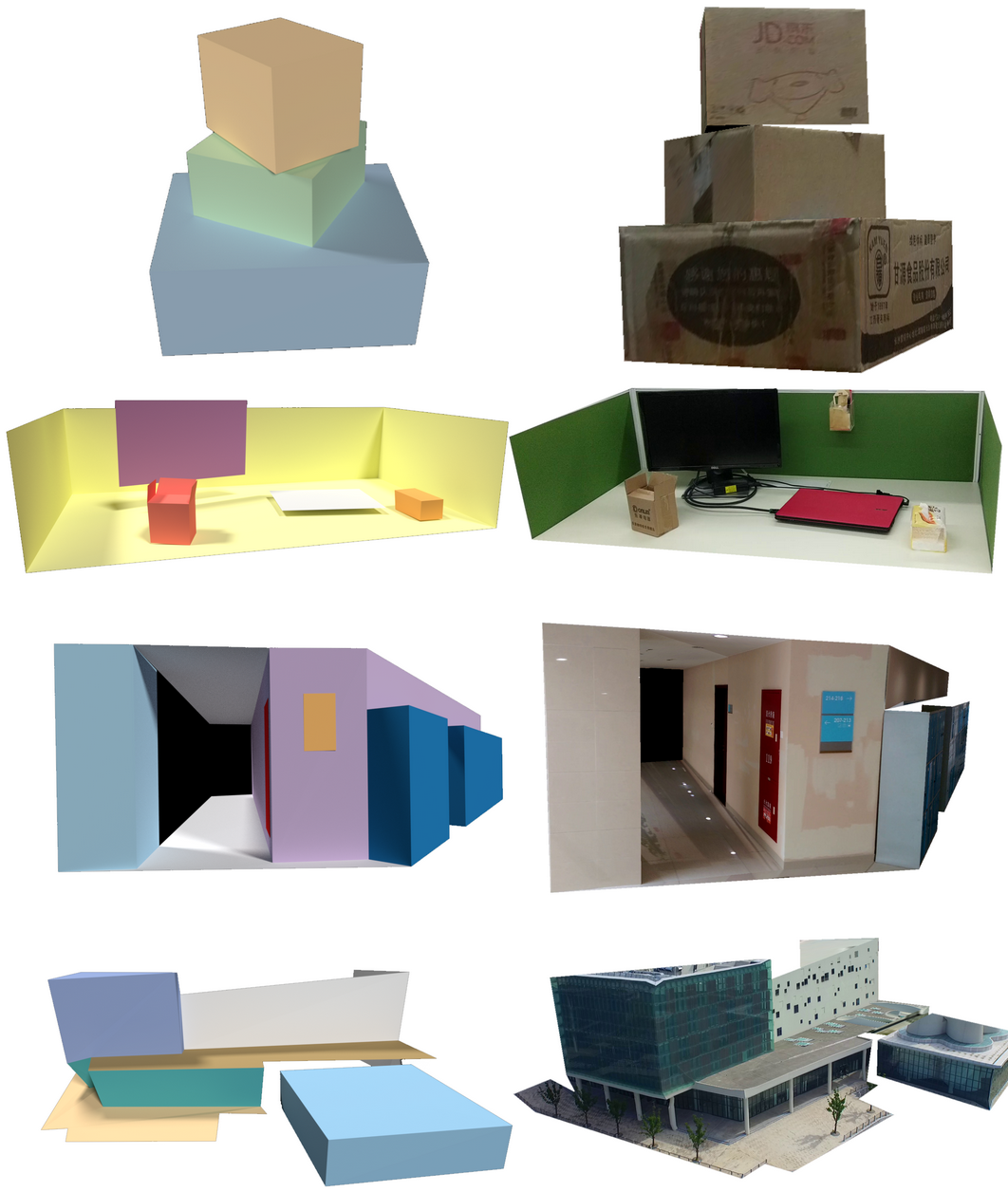}
		\caption{3D models (color shaded and textured) obtained using our algorithm on data set in Figure \ref{fig:res_example}. Our method faithfully recovers both the plane geometry and plane arrangements.}
        \label{fig:res}
	\end{figure}
		
		\begin{figure}[t]
			\centering
			\includegraphics[width=\linewidth]{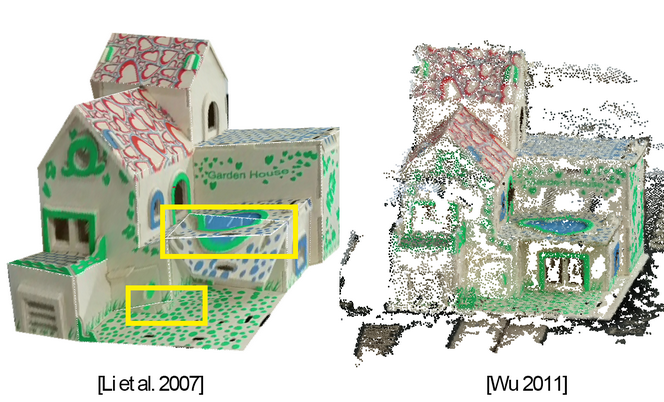}
			\caption{Results of toy house model obtained using the method of \cite{Li07x} and the method of VisualSFM (Wu 2011 \cite{Wu_visualSFM}).}
            \label{fig:compare}
		\end{figure}
		
		In this section, we experiment using real data to fully evaluate our algorithm. Figure \ref{fig:res_example} shows the example 3D models and scenes we use in our experiments. They include stacking boxes in a dorm, desktop workspace in a computer lab, corridor inside a building, a toy house, and a school library. They are captured either by a cell phone or by a UAV (the library scene). These models span a typical set of objects in manmade environments and many of them consist of a lot of planar structures which fit perfectly to our algorithm.

		
		When it comes to real data, things get a little different. The first thing is that the camera's intrinsic parameters are unknown. We solve this by assuming only the focal length of the camera is missing, and we try different values to unproject the 2d image and choose the one with the least error. The second is that real data always has blur due to the camera's unstable moving and the resolution limitation, which has negative effects on our tracking process, even if the tracking is helped by users. So providing structure constraints will help to improve the optimization result.
		
	Figure \ref{fig:res} shows the reconstruction results and Table \ref{tbl:stats} shows the statistics of the generated results. We render the reconstructed results using both shaded 3D model and textured model for clear illustration. To texture a planar face, we use a similar technology as in \cite{Sinha:2008:IAM}. Our method allows the user to manually adjust the dynamic tracking process once failures were detected (see the point adjustment statistics in Table \ref{tbl:stats}). We observe that such cases happen typically at places where occlusion happens (e.g., the occluded points in the toy house model) or when the structure lines to track blends with the background (Figure \ref{fig:pathFinding} and Figure \ref{fig:res_example} bottom left). We believe these two cases are inherently challenging to handle even with our human perception. We leave it as future work.



\textbf{Time complexity.} Our algorithm consists of a tracking part and a bundle optimization part. The tracking employed a shortest path dynamic programming whose time complexity is $O(kN^2)$ with $N$ the largest window size and $k$ the number of frames (note that here our graph is layered,thus the complexity is different from a traditional all-pairs shortest path algorithm on a graph $G$ which is known to have an approximate complexity of $O(|V(G)|^3)$ . The bundle adjustment optimization runs at the same rate as traditional Sfm methods which is super fast in our case, as our input is a sparse set of plane points. It takes less than 1 minute to optimize the toy house model on a laptop with 3.2GHz CPU and 8GB RAM.

\begin{table}[t]
\begin{tabular}{l | c | c | c | c}
 & frames & planes & points (adj.) & time(s)\\
\hline
boxes & 300 & 11 & 23(70) & 380 \\
hall & 350 & 12 & 38(100) & 420 \\
toy house & 300 & 18 & 35(40) & 270 \\
desktop & 150 & 14 & 36(5) & 150 \\
library & 500 & 18 & 41(50) & 330
\end{tabular}
\vspace{0.1in}
\caption{\small Statistics of exemplar scenes used in our paper (Figure \ref{fig:toy_house}, \ref{fig:res_example}). The fourth column records the number of points to track and the count of user adjustment performed during tracking. The number mainly comes from points in occluded and fuzzy regions where continuous adjustments are required throughout the sequence.}
\label{tbl:stats}
\end{table}

\textbf{Comparison.} We conducted a pilot comparison with two state-of-the-art 3D reconstruction methods which we consider to be relevant. The first one is the planar reconstruction method proposed by Li et al. \cite{Li07x} and the other is the famous structure from motion software VisualSFM proposed by Wu et al. \cite{Wu_visualSFM,Wu_towardslinear-time}. Figure \ref{fig:compare} shows their results. Without any structural constraints, the VisualSFm system merely generated a set of incomplete point cloud while the loose symmetry and coplanar constraints used in the method of \cite{Li07x} still cannot guarantee a convergence of the output to a desired one especially when only dealing with a single image (see the drifted faces). Our method faithfully recovers all planar faces and their inter-relations.
	
\textbf{Limitation.} By now, our method requires the user to specify an initial set of corner points and edges which constitute the planar faces. The initial specification typically takes around 2-5 minutes. This is the main limitation of our algorithm. By far, we are not aware of any automatic algorithms that can robustly identify planar regions from RGB images. An intriguing direction to explore is to use some deep-learning based approaches for detecting planar regions. Another limitation is that our structure-based dynamic tracking could fail at places where the edges get weak or occlusion happens. This is unavoidable due to the inherent noise and motion blur during video capture. A pre-denoise or deblurring process could alleviate the problem a bit but completely solving such problems requires more significant efforts as this needs a semantic understanding of the underlying scene.

\section{Conclusion}
	In conclusion, this paper provides a semi-automatic 3D reconstruction algorithm that recovers a set of structured planes along with their arrangements from a video. Our key contribution is a structure model represented as a set of planes whose arrangements form a faithful description of the scene model. We propose a dynamic point tracking algorithm which explicitly exploits the structure lines as effective means for identifying reliable corner point locations. Besides, a structure-augmented optimization framework with bundle adjustment is introduced to jointly optimize the plane arrangements and the plane geometry. Our future work will consider to combine the traditional Sfm process with automatic structure analysis to enable a fully automated 3D reconstruction pipeline, which we believe will open up new possibilities in the area of structure-based 3D reconstruction and bring potential influence to the community.

{\small
\bibliographystyle{ieee}
\bibliography{egbib}

\begin{thebibliography}{10}\itemsep=-1pt

\bibitem{bailer2015flow}
C.~Bailer, B.~Taetz, and D.~Stricker.
\newblock Flow fields: Dense correspondence fields for highly accurate large
  displacement optical flow estimation.
\newblock In {\em Proceedings of the IEEE International Conference on Computer
  Vision}, pages 4015--4023, 2015.

\bibitem{brox2011large}
T.~Brox and J.~Malik.
\newblock Large displacement optical flow: descriptor matching in variational
  motion estimation.
\newblock {\em IEEE transactions on pattern analysis and machine intelligence},
  33(3):500--513, 2011.

\bibitem{Ceylan:2014:CSS}
D.~Ceylan, N.~J. Mitra, Y.~Zheng, and M.~Pauly.
\newblock Coupled structure-from-motion and 3d symmetry detection for urban
  facades.
\newblock {\em ACM Trans. Graph.}, 33(1):2:1--2:15, 2014.

\bibitem{cheng1995mean}
Y.~Cheng.
\newblock Mean shift, mode seeking, and clustering.
\newblock {\em IEEE transactions on pattern analysis and machine intelligence},
  17(8):790--799, 1995.

\bibitem{furukawa2009manhattan}
Y.~Furukawa, B.~Curless, S.~M. Seitz, and R.~Szeliski.
\newblock Manhattan-world stereo.
\newblock In {\em Computer Vision and Pattern Recognition, 2009. CVPR 2009.
  IEEE Conference on}, pages 1422--1429. IEEE, 2009.

\bibitem{furukawa2010towards}
Y.~Furukawa, B.~Curless, S.~M. Seitz, and R.~Szeliski.
\newblock Towards internet-scale multi-view stereo.
\newblock In {\em Computer Vision and Pattern Recognition (CVPR), 2010 IEEE
  Conference on}, pages 1434--1441. IEEE, 2010.

\bibitem{furukawa2010accurate}
Y.~Furukawa and J.~Ponce.
\newblock Accurate, dense, and robust multiview stereopsis.
\newblock {\em IEEE transactions on pattern analysis and machine intelligence},
  32(8):1362--1376, 2010.

\bibitem{hare2012efficient}
S.~Hare, A.~Saffari, and P.~H. Torr.
\newblock Efficient online structured output learning for keypoint-based object
  tracking.
\newblock In {\em Computer Vision and Pattern Recognition (CVPR), 2012 IEEE
  Conference on}, pages 1894--1901. IEEE, 2012.

\bibitem{hartley2003multiple}
R.~Hartley and A.~Zisserman.
\newblock {\em Multiple view geometry in computer vision}.
\newblock Cambridge university press, 2003.

\bibitem{horn1981determining}
B.~K. Horn and B.~G. Schunck.
\newblock Determining optical flow.
\newblock {\em Artificial intelligence}, 17(1-3):185--203, 1981.

\bibitem{ikehata2015structured}
S.~Ikehata, H.~Yang, and Y.~Furukawa.
\newblock Structured indoor modeling.
\newblock In {\em Proceedings of the IEEE International Conference on Computer
  Vision}, pages 1323--1331, 2015.

\bibitem{izadi2011kinectfusion}
S.~Izadi, D.~Kim, O.~Hilliges, D.~Molyneaux, R.~Newcombe, P.~Kohli, J.~Shotton,
  S.~Hodges, D.~Freeman, A.~Davison, et~al.
\newblock Kinectfusion: Real-time 3d reconstruction and interaction using a
  moving depth camera.
\newblock In {\em Proceedings of the 24th annual ACM symposium on User
  interface software and technology}, pages 559--568. ACM, 2011.

\bibitem{jia-3d-stability-pami}
Z.~Jia, A.~Gallagher, A.~Saxena, and T.~Chen.
\newblock 3d reasoning from blocks to stability.
\newblock {\em IEEE Trans PAMI}, 37(5):905--918, 2015.

\bibitem{jiang2009symmetric}
N.~Jiang, P.~Tan, and L.-F. Cheong.
\newblock Symmetric architecture modeling with a single image.
\newblock In {\em ACM Transactions on Graphics (TOG)}, volume~28, page 113.
  ACM, 2009.

\bibitem{lempitsky2008fusionflow}
V.~Lempitsky, S.~Roth, and C.~Rother.
\newblock Fusionflow: Discrete-continuous optimization for optical flow
  estimation.
\newblock In {\em Computer Vision and Pattern Recognition, 2008. CVPR 2008.
  IEEE Conference on}, pages 1--8. IEEE, 2008.

\bibitem{li_globFit_sigg11}
Y.~Li, X.~Wu, Y.~Chrysanthou, A.~Sharf, D.~Cohen-Or, and N.~J. Mitra.
\newblock Globfit: Consistently fitting primitives by discovering global
  relations.
\newblock {\em ACM Transactions on Graphics}, 30(4):52:1--52:12, 2011.

\bibitem{Li07x}
Z.~Li, J.~Liu, and X.~Tang.
\newblock A closed-form solution to 3d reconstruction of piecewise planar
  objects from single images.
\newblock In {\em 2007 IEEE Conference on Computer Vision and Pattern
  Recognition}, pages 1--6. IEEE, 2007.

\bibitem{lucas1981iterative}
B.~D. Lucas, T.~Kanade, et~al.
\newblock An iterative image registration technique with an application to
  stereo vision.
\newblock In {\em IJCAI}, volume~81, pages 674--679, 1981.

\bibitem{ma2012invitation}
Y.~Ma, S.~Soatto, J.~Kosecka, and S.~S. Sastry.
\newblock {\em An invitation to 3-d vision: from images to geometric models},
  volume~26.
\newblock Springer Science \& Business Media, 2012.

\bibitem{MonszpartEtAl:RAPter:2015}
A.~Monszpart, N.~Mellado, G.~Brostow, and N.~Mitra.
\newblock {RAP}ter: Rebuilding man-made scenes with regular arrangements of
  planes.
\newblock {\em {ACM SIGGRAPH 2015}}, 34(4):103:1--103:12, 2015.

\bibitem{Mura:2015:ARD}
C.~Mura, O.~Mattausch, A.~{Jaspe Villanueva}, E.~Gobbetti, and R.~Pajarola.
\newblock Automatic room detection and reconstruction in cluttered indoor
  environments with complex room layouts.
\newblock In {\em Proc. Spring conference on Computer Graphics - Invited CAG
  Presentation}, 2015.
\newblock To appear.

\bibitem{ondruvska2015mobilefusion}
P.~Ondr{\'u}{\v{s}}ka, P.~Kohli, and S.~Izadi.
\newblock Mobilefusion: Real-time volumetric surface reconstruction and dense
  tracking on mobile phones.
\newblock {\em IEEE transactions on visualization and computer graphics},
  21(11):1251--1258, 2015.

\bibitem{Seitz06acomparison}
S.~M. Seitz, B.~Curless, J.~Diebel, D.~Scharstein, and R.~Szeliski.
\newblock A comparison and evaluation of multi-view stereo reconstruction
  algorithms.
\newblock In {\em CVPR}, pages 519--528, 2006.

\bibitem{Sinha:2008:IAM}
S.~N. Sinha, D.~Steedly, R.~Szeliski, M.~Agrawala, and M.~Pollefeys.
\newblock Interactive 3d architectural modeling from unordered photo
  collections.
\newblock {\em ACM Trans. Graph.}, 27(5):159:1--159:10, Dec. 2008.

\bibitem{snavely2006photo}
N.~Snavely, S.~M. Seitz, and R.~Szeliski.
\newblock Photo tourism: exploring photo collections in 3d.
\newblock In {\em ACM transactions on graphics (TOG)}, volume~25, pages
  835--846. ACM, 2006.

\bibitem{sturm1999method}
P.~Sturm and S.~Maybank.
\newblock A method for interactive 3d reconstruction of piecewise planar
  objects from single images.
\newblock In {\em The 10th British machine vision conference (BMVC'99)}, pages
  265--274. The British Machine Vision Association (BMVA), 1999.

\bibitem{vandenHengel:2007}
A.~van~den Hengel, A.~Dick, T.~Thorm\"{a}hlen, B.~Ward, and P.~H.~S. Torr.
\newblock Videotrace: Rapid interactive scene modelling from video.
\newblock In {\em ACM SIGGRAPH 2007 Papers}, SIGGRAPH '07, New York, NY, USA,
  2007. ACM.

\bibitem{wedel2009structure}
A.~Wedel, D.~Cremers, T.~Pock, and H.~Bischof.
\newblock Structure-and motion-adaptive regularization for high accuracy optic
  flow.
\newblock In {\em ICCV}, pages 1663--1668, 2009.

\bibitem{weinzaepfel:hal}
P.~Weinzaepfel, J.~Revaud, Z.~Harchaoui, and C.~Schmid.
\newblock {DeepFlow: Large displacement optical flow with deep matching}.
\newblock In {\em {ICCV 2013 - IEEE International Conference on Computer
  Vision}}, pages 1385--1392, Sydney, Australia, Dec. 2013. {IEEE}.

\bibitem{Wu_visualSFM}
C.~Wu.
\newblock Visualsfm: A visual structure from motion system, 2011.

\bibitem{Wu_towardslinear-time}
C.~Wu.
\newblock Towards linear-time incremental structure from motion, 2013.

\bibitem{Xiao:2014:RWM}
J.~Xiao and Y.~Furukawa.
\newblock Reconstructing the world's museums.
\newblock {\em Int. J. Comput. Vision}, 110(3):243--258, Dec. 2014.

\bibitem{xiong2013automatic}
X.~Xiong, A.~Adan, B.~Akinci, and D.~Huber.
\newblock Automatic creation of semantically rich 3d building models from laser
  scanner data.
\newblock {\em Automation in Construction}, 31:325--337, 2013.

\bibitem{xu2016SA}
M.~Xu, M.~Li, W.~Xu, Z.~Deng, Y.~Yang, , and K.~Zhou.
\newblock Interactive mechanism modeling from multi-view images.
\newblock In {\em ACM Transactions on Graphics (TOG)}, volume~35. ACM, 2016.

\bibitem{Xue2011}
T.~Xue, J.~Liu, and X.~Tang.
\newblock Symmetric piecewise planar object reconstruction from a single image.
\newblock In {\em Computer Vision and Pattern Recognition (CVPR), 2011 IEEE
  Conference on}, June 2011.

\bibitem{xue2011symmetric}
T.~Xue, J.~Liu, and X.~Tang.
\newblock Symmetric piecewise planar object reconstruction from a single image.
\newblock In {\em Computer Vision and Pattern Recognition (CVPR), 2011 IEEE
  Conference on}, pages 2577--2584. IEEE, 2011.

\end{thebibliography}
}

\end{document}